\documentclass[letterpaper, 10 pt, conference]{ieeeconf}  

\IEEEoverridecommandlockouts                              
\usepackage{amsmath} 
\usepackage{algorithm}
\usepackage{algorithmic}
\usepackage{graphicx}
\usepackage{tikz}
\usepackage{booktabs}
\usepackage{multirow}
\usetikzlibrary{arrows}
\usetikzlibrary{arrows.meta}
\tikzset{>={Latex[width=2mm,length=2mm]}}
\usetikzlibrary {patterns,shapes,shadows,decorations.pathreplacing}
\usepackage{multicol}
\title{\LARGE \bf
Resource-Constrained Simultaneous Detection and Labeling\\ of Objects in High-Resolution Satellite Images
}

\author{\large Gilbert Rotich$^1$, Rodrigo Minetto$^2$ and Sudeep Sarkar$^1$\\ 
$^1$ University of South Florida (USF)\\
$^2$ The Federal University of Technology - Paran\'{a} (UTFPR)\\
{\tt rminetto@dainf.ct.utfpr.edu.br}
}

\begin{document}

\maketitle
\thispagestyle{empty}
\pagestyle{empty}

\begin{abstract}
We describe a strategy for detection and classification of man-made objects in large high-resolution satellite photos under computational resource constraints. We detect and classify candidate objects by using five pipelines of convolutional neural network processing (CNN), run in parallel. Each pipeline has its own unique strategy for fine tunning parameters, proposal region filtering, and dealing with image scales. The conflicting region proposals are merged based on region confidence and not just based on overlap areas, which improves the quality of the final bounding-box regions selected. We demonstrate this strategy using the recent xView challenge, which is a complex benchmark with more than 1,100 high-resolution images, spanning 800,000 aerial objects around the world covering a total area of 1,400 square kilometers at 0.3 meter ground sample distance. To tackle the resource-constrained problem posed by the xView challenge, where inferences are restricted to be on CPU with 8GB memory limit, we used lightweight CNN's trained with the single shot detector algorithm. Our approach was competitive on sequestered sets; it was ranked third.
\end{abstract}

\section{INTRODUCTION}

The localization and classification of geographical regions in high-resolution aerial images provides critical information for analysts and police makers around the world to make decisions about territorial defense, humanitarian assistance and environmental conservation policies. Although extensively studied~\cite{Xia2018,4270421,4587359,bastani2018roadtracer}, such research problems are still very challenging. The technical difficulties have been exposed by the recent large challenge problems that have been constructed such as the Spacenet challenges~\cite{van2018spacenet} (https://spacenetchallenge.github.io/), IARPA Functional Map of the World (fMoW) challenge~\cite{christie2018functional}, and the DoD xView challenge~\cite{xview2018}.

SpaceNet challenge focused on the problem of building and road networks in satellite images of five metropolitan areas with over 685,000 footprints. The fMoW contest published  one of the hardest and largest benchmark for region classification in aerial images to date, with 1 million images around 100,000 globe locations. The goal in fMoW was the classification of a given region as one of 62 target classes or as a false detection. The xView challenge published another complex benchmark with more than 1,100 high-resolution images, spanning a total area of 1,400 square kilometers collected at 0.3 meter ground sample distance (GSD), and containing more than 800,000 aerial objects around the world. The goal in xView was the localization and classification of such objects into 60 classes. 

These novel benchmarks in remote sensing foster major breakthroughs in machine learning, by addressing complex problems such as 
\begin{itemize}
  \item \textbf{Fine grained categorization}, which is the the classification of  visually-similar objects from subordinate categories. For instance, in the xView challenge for truck vehicles there are eight different sub-categories: pickup truck, utility truck, cargo truck, truck with box, truck tractor trailer, truck with flatbed, truck with liquid.
  \item \textbf{Resource-constrained learning} by imposing limitation on computational resources for inference and learning. This is necessary, for instance, if the final deployment is in unmanned aerial vehicle (drone), where code has to run efficiently on embedded low-power processors.  
  \item \textbf{Class imbalance}, which is the problem of learning from an 
 unequal number of observations per class. This is an inherent problem in remote sensing, where, we usually have many instances of common objects like cars, buses or buildings and few instances of other objects like excavators, locomotives and helipads.  
  \item \textbf{Spatial learning}, which is the detection and classification of objects embedded in clutter background, with large scale variation and with partial occlusion by clouds or shadows. These problems result in high intraclass variations and considerable interclass confusion.
  \item \textbf{Temporal learning}, which is the problem of learning from image sequences of the same geographical scene, recorded in arbitrary satellite viewpoints and time periods. For instance, detecting such changes could help in damage assessment and rescue efforts in case of natural disasters or in environmental conversation in case of deforestation.
 \end{itemize}

\begin{figure}[!htb]
\centering
\begin{tikzpicture}

  \draw(0.0,6.1) node[inner sep=0pt] (img1) {
     \includegraphics[width=8.0cm,height=5.8cm]{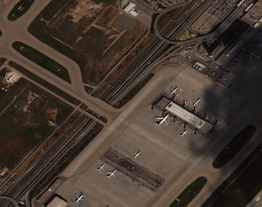}   
  };
  
  \draw(0.0,0.0) node[inner sep=0pt, opacity=1.0] (img1) {
     \includegraphics[width=8.0cm,height=5.8cm]{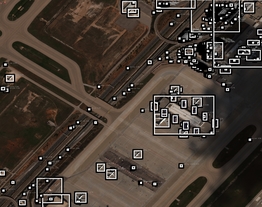}   
  };

  \draw(-2.2,0.8) node[inner sep=0pt] (img1) {        \includegraphics[width=3.3cm,height=3.3cm]{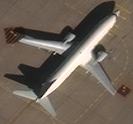}     };
   
  \path[->](-2.2,-0.1) node[black,very thick,fill=white,fill opacity=0.9, inner sep=1pt]  {\small \textbf{c = 2 ({\scriptsize passenger-plane})}};
  
  \path[->](-2.2,-0.55) node[black,very thick,fill=white,fill opacity=0.9, inner sep=1pt]  {\small \textbf{w = 0.9 ({\scriptsize confidence})}};
  
  \path[->](-3.3,2.6) node[black,very thick,fill=white,fill opacity=0.9, inner sep=1pt]  {\small \textbf{($x_1,y_1$)}};
  
    \path[->](-1.1,-1.0) node[black,very thick,fill=white,fill opacity=0.9, inner sep=1pt]  {\small \textbf{($x_2,y_2$)}};
    
  \draw[white=0.9,very thick] (-3.8,2.4) rectangle (-0.6,-0.8);
  \draw[black,fill=white] (-3.8,2.4) circle (.5ex);
  \draw[black,fill=white] (-0.6,-0.8) circle (.5ex);
  
  \draw[white,very thick, dotted] (-0.8,-1.7) -- (-0.6,-0.8);
  \draw[white,very thick, dotted] (-1.0,-1.7) -- (-3.8,-0.8);
  
\end{tikzpicture}
  \caption{xView challenge: (top) the input data for the problem is a digital aerial image; (bottom) the output is a list of regions, where each region $r_i$ is defined by three attributes: a rectangular axis-aligned box $b = (x_1, y_1, x_2, y_2)$; the object category $c$; and a classification score $w$. For the sake of simplicity, we assume that the object categories are mapped into the interval [0, N].} 
  \label{fig:xview-challenge}
\end{figure}
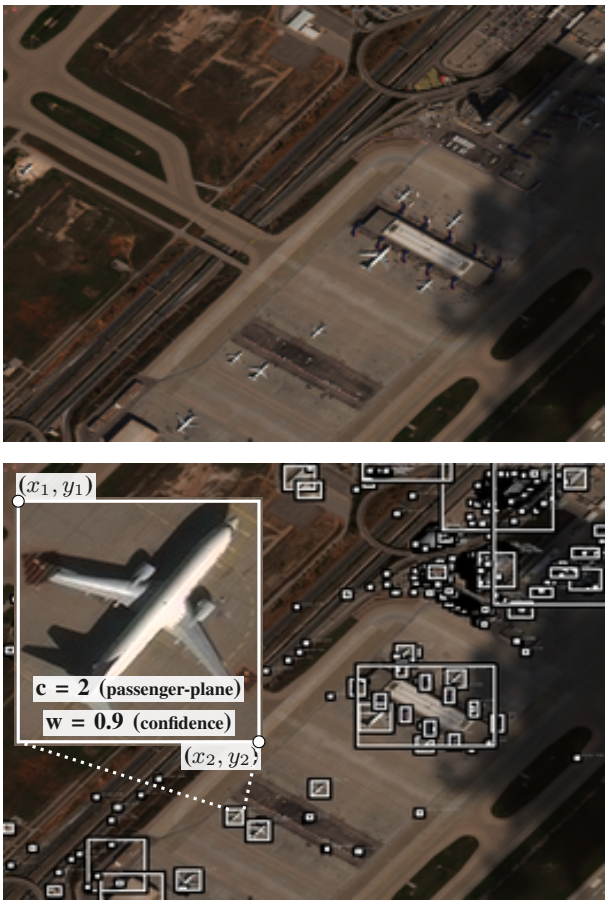

In this paper we describe a framework for simultaneous detection and classification of  objects in high-resolution aerial images. 
The input data for this problem is an aerial image, see Figure~\ref{fig:xview-challenge} (top), and the output consists of $n$ regions
\begin{equation}
  \mathcal{R} = \{r_1, r_2, \dots, r_n\}
\end{equation}
where each region $r_i$ is defined by an axis-aligned rectangular box ${r_i}(b) = (x_1, y_1, x_2, y_2)$ where $(x_1,y_1)$ and $(x_2,y_2)$ represent the upper left and bottom right corner's, respectively; an integer ${r_i}(c) \in \{1,2,\dots,N\}$ that represent the object category; and a confidence score ${r_i}(w)$, for the classification, expressed as a real number, within the interval $[0,1]$. An example of an output is shown in Figure~\ref{fig:xview-challenge} (bottom). 

We detect and classify candidate regions by using five pipelines of convolutional neural networks (CNN) to cope with the aforementioned problems in remote sensing (fine grained categorization, class imbalance, etc). 

Each pipeline has its own unique strategy for fine tunning parameters, proposal region filtering, and dealing with image scales. The conflicting region proposals are merged based on region confidence and not just based on overlap areas, which improves the quality of the final bounding-box regions selected. To tackle the resource-constrained problem, we used lightweight CNN's trained with the single shot detector algorithm as the core deep learning approach.

The remainder of the paper is organized as follows.
The proposed framework is described in Section~\ref{sec:framework}, and its experimental evaluation is reported in Section~\ref{sec:results}. We analyze the results and discuss research extensions in Sections~\ref{sec:discussion} and~\ref{sec:future_work}.

\section{Proposed Framework}~\label{sec:framework}

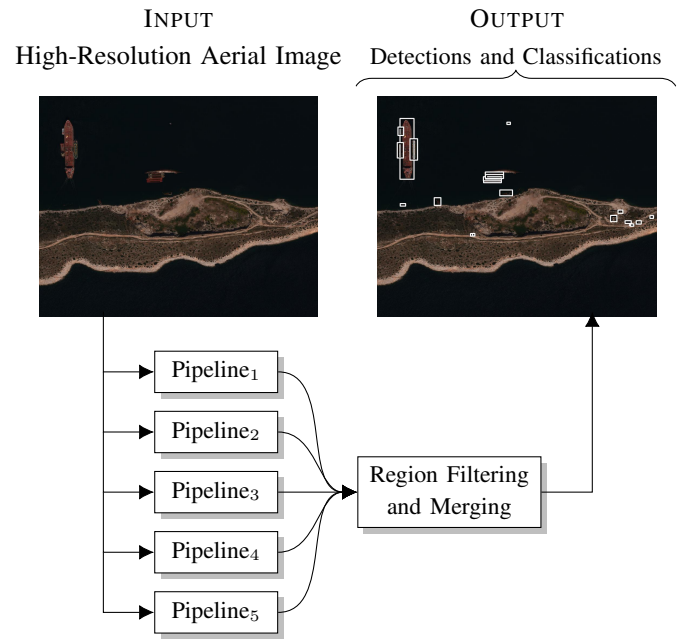
\begin{figure}[!htb]
 \centering
 \begin{tikzpicture}
   \tikzset{blockr/.style={draw, rectangle, text centered, drop shadow, fill=white, text width=1.4cm, minimum height=0.55cm}}
   \tikzset{blockt/.style={draw, rectangle, text centered, drop shadow, fill=white, text width=2.2cm, minimum height=0.55cm}}
    \tikzset{blockc/.style={draw, circle, fill=black, inner sep=1.5pt}}
     \tikzset{blockd/.style={draw, circle, color=red, fill=red, inner sep=1.5pt}}
    
   \draw(-0.5,5.4) node[text centered, inner sep=0pt] (input) {
      \includegraphics[width=3.7cm]{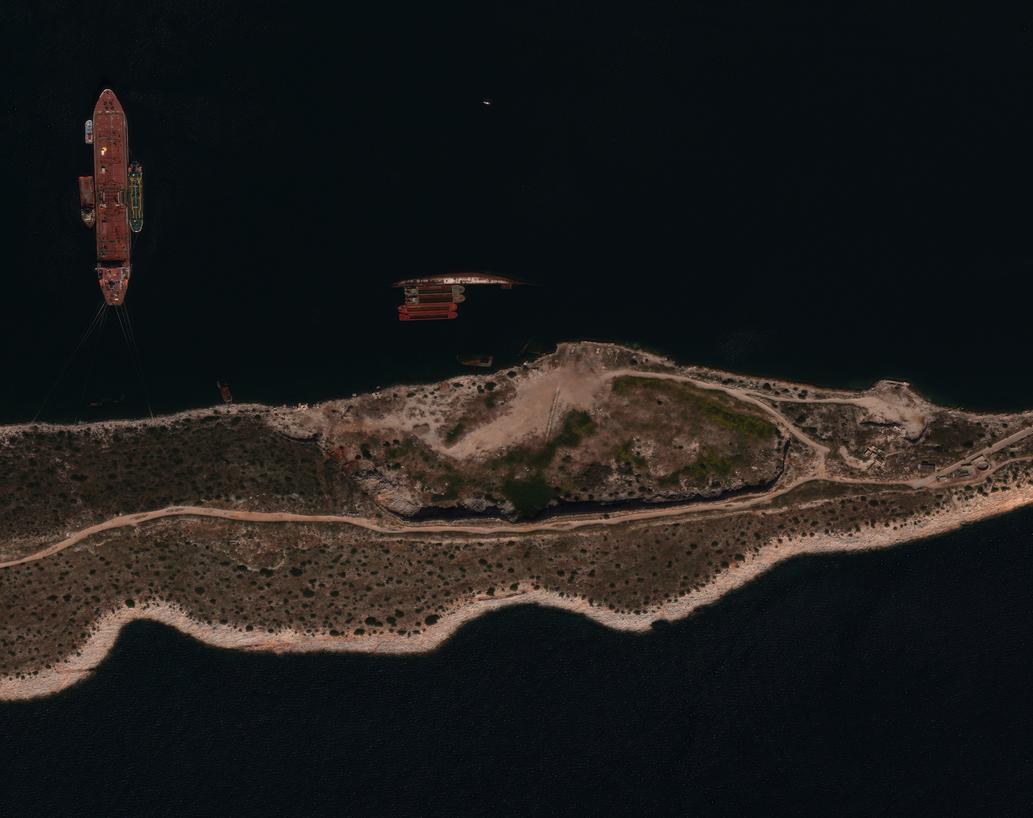}   };
      
  \draw(+4.0,5.4) node[text centered, inner sep=0pt] (output) {
      \includegraphics[width=3.7cm]{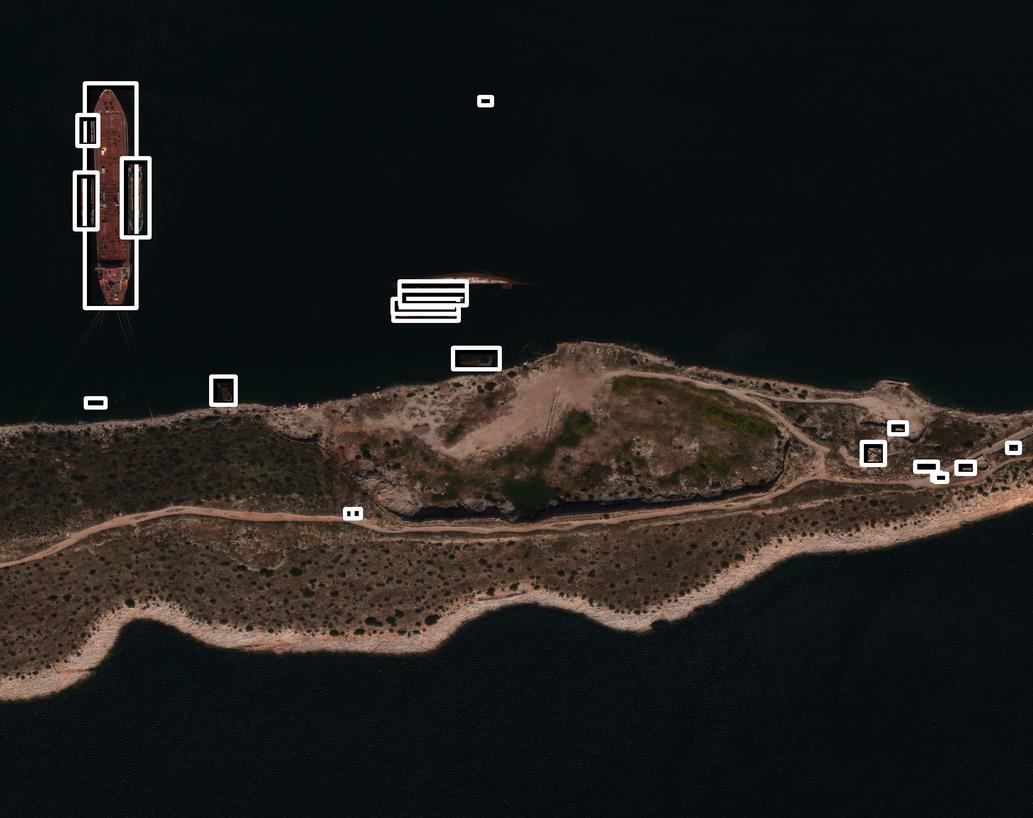}   };     
    
   \path[->](0.0,3.2) node[blockr] (r3) {
      \small Pipeline$_1$
   };
   
   \path[->](0.0,2.4) node[blockr] (r4) {
      \small Pipeline$_2$
   };
   
   \path[->](0.0,1.6) node[blockr] (r5) {
      \small Pipeline$_3$
   };
   
   \path[->](0.0,0.8) node[blockr] (r6) {
      \small Pipeline$_4$
   };

   \path[->](0.0,0.0) node[blockr] (r7) {
      \small Pipeline$_5$
   };

   \draw[->] (-1.5,4) |- (r3.west);
   \draw[->] (-1.5,3.2) |- (r4.west);
   \draw[->] (-1.5,2.4) |- (r5.west);
   \draw[->] (-1.5,1.6) |- (r6.west);
   \draw[->] (-1.5,0.8) |- (r7.west);

   \path[->](3.1,1.6) node[blockt] (nms) {
      \small Region Filtering and Merging
   };
   
   \draw (r3) edge[out=0,in=180,->] (nms);
   \draw (r4) edge[out=0,in=180,->] (nms);
   \draw (r5) edge[out=0,in=180,->] (nms);
   \draw (r6) edge[out=0,in=180,->] (nms);
   \draw (r7) edge[out=0,in=180,->] (nms);
    
    \draw[->] (nms) -| (5,4);
   
   \draw(-0.5,7.9) node[text centered, text width=1.8cm] (text4) {
      {\normalsize \textsc{Input}}
   };
 
 \draw(-0.5,7.38) node[text centered, text width=4.4cm,inner sep=0pt] (text4) {
      {\normalsize High-Resolution Aerial Image}
   };
   
   \draw(4.0,7.9) node[text centered, text width=1.8cm] (text4) {
      {\normalsize \textsc{Output}}
   };
     
   \draw [decorate,decoration={brace,amplitude=6pt},xshift=-4pt,yshift=0pt] (2.0,7.0) -- (6.25,7.0) node [black,midway,xshift=+0.0cm, yshift=0.4cm] 
{\small Detections and Classifications};
 
\end{tikzpicture} 
  \caption{Framework flowchart: the input image is fed to five detection and classification pipelines and the outcomes are filtered and merged.}
  \label{fig:framework}
\end{figure}

\begin{figure}[!htb]
 \begin{tikzpicture}
   \tikzset{blockr/.style={draw, rectangle, text centered, drop shadow, fill=white, text width=2.8cm, minimum height=0.55cm}}
   \tikzset{blockt/.style={draw, rectangle, text centered, drop shadow, fill=white, text width=2.9cm, minimum height=0.55cm}}
   \tikzset{blocks/.style={draw, rectangle, text centered, drop shadow, fill=white, text width=4.9cm, minimum height=0.55cm}}
    \tikzset{blockc/.style={draw, circle, fill=black, inner sep=1.5pt}}
     \tikzset{blockd/.style={draw, circle, color=red, fill=red, inner sep=1.5pt}}
    
   \draw(-0.5,10.76) node[text centered, inner sep=0pt] (input) {
      \includegraphics[width=3.6cm]{figures/2547.jpg}
   };
   
   \path[->](-0.5,8.55) node[blockr] (block1) {
      \small Image Scaling
   };
   
    \draw[->] (block1.east) -- (1.5,8.55);
      
   \draw(+4.0,10.2) node[text centered, inner sep=0pt] (scaled) {
      \includegraphics[width=5.0cm]{figures/2547.jpg}
   };     
    
   \path[->](4.0,7.45) node[blockr] (block2) {
      \small Image Splitting
   };
   
   \path[->](2.0,3.1) node[blockt] (cnn) {
      \small CNN Inference
   };
   
   \path[->](2.0,-1.0) node[blockr] (rescale) {
      \small Region Rescaling
   };
   
   \draw(-1.2,5.3) node[text centered, inner sep=0pt] (split0) {
      \includegraphics[width=1.8cm,height=1.8cm]{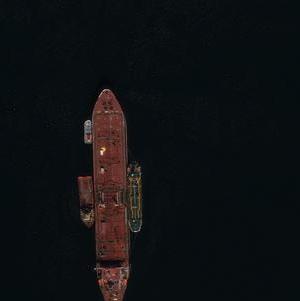}
   };  
   
   \draw(+0.8,5.3) node[text centered, inner sep=0pt] (split1) {
      \includegraphics[width=1.8cm,height=1.8cm]{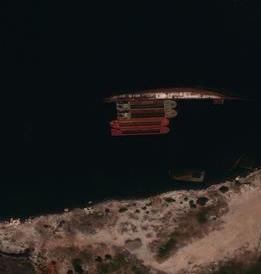}
   };  
   
   \draw(+2.8,5.3) node[text centered, inner sep=0pt] (split2) {
      \includegraphics[width=1.8cm,height=1.8cm]{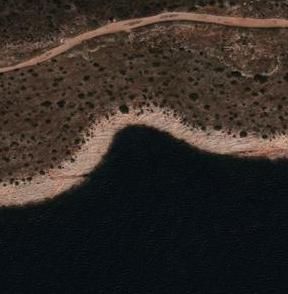}
   };  
   
    \draw(+4.2,5.3) node[text centered, text width=1.3cm] (dots1) {
      {\large $\dots$}
   };
   
   \draw(+5.5,5.3) node[text centered, inner sep=0pt] (split3) {
      \includegraphics[width=1.8cm,height=1.8cm]{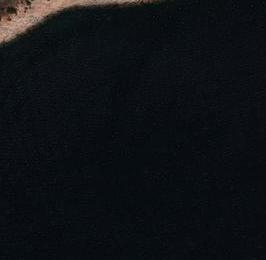}
   };

  \draw[->] (input) -- (block1);
  \draw[->] (scaled) -- (block2);
   
 \draw(+4.0,6.7) node[inner sep=-2pt] (anchor) {};  
   
 \draw[] (block2) -- (anchor);  
 \draw[->] (+0.8,6.7) -| (split0);
 \draw[->] (+2.8,6.7) -| (split1);
 \draw[->] (anchor) -| (split2);
 \draw[->] (anchor) -| (split3);
 
  \draw(+2.0,3.9) node[inner sep=-2pt] (anchor2) {};

 \draw[] (split0) |- (anchor2);
 \draw[] (split1) |- (anchor2);
 \draw[] (split2) |- (anchor2);
 \draw[] (split3) |- (anchor2);
 \draw[->] (anchor2) -- (cnn);

   \draw(-1.2,1.0) node[text centered, inner sep=0pt] (det0) {
      \includegraphics[width=1.8cm,height=1.8cm]{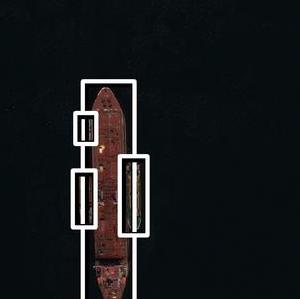}
   };  
   
   \draw(+0.8,1.0) node[text centered, inner sep=0pt] (det1) {
      \includegraphics[width=1.8cm,height=1.8cm]{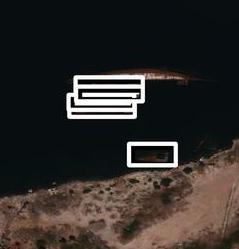}
   };  
   
   \draw(+2.8,1.0) node[text centered, inner sep=0pt] (det2) {
      \includegraphics[width=1.8cm,height=1.8cm]{figures/2547_crop3.jpg}
   };  
   
    \draw(+4.2,1.0) node[text centered, text width=1.3cm] (dots1) {
      {\large $\dots$}
   };
   
   \draw(+5.5,1.0) node[text centered, inner sep=0pt] (det3) {
      \includegraphics[width=1.8cm,height=1.8cm]{figures/2547_crop4.jpg}
   };  
  
    \draw(+2.0,2.4) node[inner sep=-2pt] (anchor3) {};  
   
 \draw[-] (cnn) -- (anchor3);
 \draw[->] (anchor3) -| (det0);
 \draw[->] (anchor3) -| (det1);
 \draw[->] (anchor3) -| (det2);
 \draw[->] (anchor3) -| (det3);
 
  \draw(+2.0,-0.3) node[inner sep=-2pt] (anchor4) {}; 
  
 \draw[] (det0) |- (anchor4);
 \draw[] (det1) |- (anchor4);
 \draw[] (det2) |- (anchor4);
 \draw[] (det3) |- (anchor4);
 \draw[->] (anchor4) -- (rescale);

 \draw(2.0,-2.0) node[text centered, text width=2.8cm] (text4) {
      {\normalsize Candidate regions }
   };
   
    \draw[->] (rescale) -- (text4);
 
\end{tikzpicture} 
  \caption{Pipeline flowchart: the input is a high-resolution image and the output is a set of regions detected by the CNN at all image slices.}
  \label{fig:pipeline_flowchart}
\end{figure}

As shown in Figure~\ref{fig:framework}, our framework consists in five pipelines for detection and labeling of objects in high-resolution aerial images. The task of the region filtering and merging module is to examine the pipeline candidate regions to discard the false detections and merge the remaining regions. The pipeline, see Figure~\ref{fig:pipeline_flowchart}, 
resize and split the original image for the inference stage. Then, the coordinates of the candidate regions are rescaled to lie within the image domain boundaries. In the rest of this section we provide a detailed description of these steps.

\subsection{Image Rescale} 

\begin{figure}[!htb]
   \centering
   \includegraphics[scale=0.7]{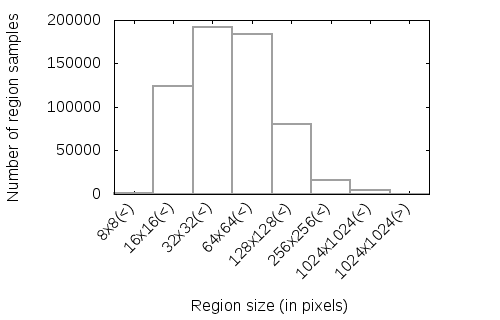}
   \caption{Object size distribution for the xView benchmark.}
   \label{fig:size_distribution} 
\end{figure}

Algorithms for region detection of high-resolution aerial images need to deal with the huge scale variation of the target objects. For instance, as shown in Figure~\ref{fig:size_distribution}, in the xView benchmark the size of the objects range from $4 \times 4$ to $3299 \times 3132$ pixels. 

Therefore, we rescaled the input image based into three classes of object size:
\begin{itemize}
   \item small objects
   \item medium objects
   \item large objects
\end{itemize}   
Specifically, we downscaled the image to detect some instances of medium and large objects and upscaled the image to detect some instances of medium and small objects. We also used the original scale to detect the three classes. The parameter settings is detailed in Section~\ref{sec:settings}. 

\subsection{Image Splitting} 

\begin{figure}[!htb]
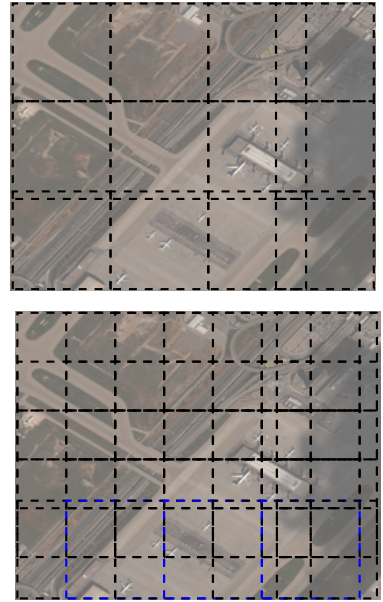

  \centering  
  \begin{tikzpicture}
   \draw(2.4,2.0) node[inner sep=0pt,opacity=0.6] (img1) {
     \includegraphics[scale=0.7]{figures/203.jpg}
   };
 
   \draw[black=0.9,thick,dashed] (0.0,0.1) rectangle (1.3,1.4);
   \draw[black=0.9,thick,dashed] (+1.3,0.1) rectangle (2.6,1.4);
   \draw[black=0.9,thick,dashed] (+2.6,0.1) rectangle (3.9,1.4); 
   \draw[black=0.9,thick,dashed] (+3.5,0.1) rectangle (4.8,1.4); 
   
   \draw[black=0.9,thick,dashed] (0.0,1.3) rectangle (1.3,2.6);
   \draw[black=0.9,thick,dashed] (+1.3,1.3) rectangle (2.6,2.6);
   \draw[black=0.9,thick,dashed] (+2.6,1.3) rectangle (3.9,2.6); 
   \draw[black=0.9,thick,dashed] (+3.5,1.3) rectangle (4.8,2.6); 
   
  \draw[black=0.9,thick,dashed] (0.0,2.6) rectangle (1.3,3.9);
   \draw[black=0.9,thick,dashed] (+1.3,2.6) rectangle (2.6,3.9);
   \draw[black=0.9,thick,dashed] (+2.6,2.6) rectangle (3.9,3.9); 
   \draw[black=0.9,thick,dashed] (+3.5,2.6) rectangle (4.8,3.9); 
  \end{tikzpicture} \vspace{8pt}
  
  \begin{tikzpicture}
  \draw(2.4,2.0) node[inner sep=0pt,opacity=0.6] (img1) {
     \includegraphics[scale=0.7]{figures/203.jpg}
   };
 
   \draw[black=0.9,thick,dashed] (0.0,2.6) rectangle (1.3,3.9);
   \draw[black=0.9,thick,dashed] (+0.65,2.6) rectangle (1.95,3.9);
   \draw[black=0.9,thick,dashed] (+1.3,2.6) rectangle (2.6,3.9); 
   \draw[black=0.9,thick,dashed] (+1.95,2.6) rectangle (3.25,3.9); 
  \draw[black=0.9,thick,dashed] (+2.6,2.6) rectangle (3.9,3.9); 
  \draw[black=0.9,thick,dashed] (+3.25,2.6) rectangle (4.55,3.9); 
  \draw[black=0.9,thick,dashed] (+3.45,2.6) rectangle (4.78,3.9); 
  
   \draw[black=0.9,thick,dashed] (0.0,1.95) rectangle (1.3,3.25);
   \draw[black=0.9,thick,dashed] (+0.65,1.95) rectangle (1.95,3.25);
   \draw[black=0.9,thick,dashed] (+1.3,1.95) rectangle (2.6,3.25); 
   \draw[black=0.9,thick,dashed] (+1.95,1.95) rectangle (3.25,3.25); 
  \draw[black=0.9,thick,dashed] (+2.6,1.95) rectangle (3.9,3.25); 
  \draw[black=0.9,thick,dashed] (+3.25,1.95) rectangle (4.55,3.25); 
  \draw[black=0.9,thick,dashed] (+3.45,1.95) rectangle (4.78,3.25); 
  
   \draw[black=0.9,thick,dashed] (0.0,1.3) rectangle (1.3,2.60);
   \draw[black=0.9,thick,dashed] (+0.65,1.3) rectangle (1.95,2.60);
   \draw[black=0.9,thick,dashed] (+1.3,1.3) rectangle (2.6,2.60); 
   \draw[black=0.9,thick,dashed] (+1.95,1.3) rectangle (3.25,2.60); 
  \draw[black=0.9,thick,dashed] (+2.6,1.3) rectangle (3.9,2.60); 
  \draw[black=0.9,thick,dashed] (+3.25,1.3) rectangle (4.55,2.60); 
  \draw[black=0.9,thick,dashed] (+3.45,1.3) rectangle (4.78,2.60); 
  
   \draw[black=0.9,thick,dashed] (0.0,0.65) rectangle (1.3,1.95);
   \draw[black=0.9,thick,dashed] (+0.65,0.65) rectangle (1.95,1.95);
   \draw[black=0.9,thick,dashed] (+1.3,0.65) rectangle (2.6,1.95); 
   \draw[black=0.9,thick,dashed] (+1.95,0.65) rectangle (3.25,1.95); 
  \draw[black=0.9,thick,dashed] (+2.6,0.65) rectangle (3.9,1.95); 
  \draw[black=0.9,thick,dashed] (+3.25,0.65) rectangle (4.55,1.95); 
  \draw[black=0.9,thick,dashed] (+3.45,0.65) rectangle (4.78,1.95); 
    
   \draw[black=0.9,thick,dashed] (0.0,0.1) rectangle (1.3,1.4);
   \draw[blue=0.9,thick,dashed] (+0.65,0.1) rectangle (1.95,1.4);
   \draw[black=0.9,thick,dashed] (+1.3,0.1) rectangle (2.6,1.4); 
   \draw[blue=0.9,thick,dashed] (+1.95,0.1) rectangle (3.25,1.4); 
  \draw[black=0.9,thick,dashed] (+2.6,0.1) rectangle (3.9,1.4); 
  \draw[blue=0.9,thick,dashed] (+3.25,0.1) rectangle (4.55,1.4); 
  \draw[black=0.9,thick,dashed] (+3.45,0.1) rectangle (4.78,1.4); 
  
  \end{tikzpicture}
   \caption{Region splitting: on top, region splitting ``without'' overlap (with the exception of the region extremes); on bottom, splitting with region overlap (50\%).}
   \label{fig:splitting}
\end{figure}

The convolutional neural network rescale the image to a fixed size for training and inference. However, the rescaling transformation may destroy the object details and compromise the detection and classification. Therefore, we split the input image into pieces with the same size and with the exact dimensions expected by the network, so as to ensure that objects will not be distorted. Furthermore, as an attempt to recover those objects that are divided by region splitting, we also use regions with overlap in two of the pipelines. See Figure~\ref{fig:splitting}. 

\subsection{Inference}

We used for inference the baseline models provided by the xView team~\cite{xview2018}. These models were trained with the Single Shot Multibox Detector (SSD)~\cite{liu2016ssd} algorithm by using 
the Inception Network~\cite{Szegedy_2015_CVPR} and two different techniques of image splitting for training: single resolution (SR) and multiple resolution (MR). For SR the original image was split into slices of $300 \times 300$ pixels, while in MR strategy they split the original image several times by using three different partitions $300 \times 300$, $400 \times 400$ and $500 \times 500$. 

We used in our pipeline a SR model fine-tuned from the baseline, with a dropout of 20\% and by using 80\% of the training samples for parameter optimization and 20\% for validation.  

\subsection{Region Rescaling}

The image scaling step changes the image domain. Therefore, it is necessary to map the coordinates of the detected regions $(\hat{x},\hat{y})$ into the coordinates $(x,y)$ of the original image. Precisely, 
\begin{eqnarray*}
  x &=& \hat{x} \times \frac{1}{\texttt{\small scale factor}} \\ \\
  y &=& \hat{y} \times \frac{1}{\texttt{\small scale factor}}
\end{eqnarray*}
where a $\texttt{\small scale factor} < 1$ downscale the image (parameter used in the image scaling step) and 
 $\texttt{\small scale factor} > 1$ upscale the image.

\subsection{Region Filtering and Merging}

The proposed framework is composed by a set of convolutional neural networks, as a consequence, some object regions may be detected multiple times. Traditionally, regions with low confidence score are filtered out by using a fixed threshold $\lambda$, and, for the remaining regions,  
a greedy \textit{non-maximum suppression} (NMS) algorithm is used to discard the region hypotheses supposed to belong to the same object. 

A standard greedy NMS algorithm, as described by Felzenszwalb~\cite{Felzenszwalb2010}, initially sorts the set of detections $\mathcal{R} = \{r_1, r_2, \dots, r_n\}$ by the confidence score. 
Then, it selects a region $r_i \in \mathcal{R}$ with highest confidence score and loops through $\mathcal{R}$, grouping other regions $r_j$ that have an \emph{intersection score} greater than a given threshold $\sigma$, that is,
\begin{equation}
 \textrm{I}({r_i}(b), {r_j}(b)) = \frac{\textrm{area\;}({r_i}(b) \cap {r_j}(b))}{\textrm{area\;}({r_i}(b))}  >  \sigma 
\end{equation}
The algorithm ends up by partitioning $\mathcal{R}$ into $k$ subsets $\mathcal{R}_1, \mathcal{R}_2, \dots \mathcal{R}_k$ of overlapping regions, and it outputs $k$ regions 
\begin{equation}
  D = \{r_1, r_2, \dots, r_k\}
\end{equation}
by selecting within each subset the region with highest confidence score.

\begin{figure}[!htb]
\centering
\begin{tikzpicture}

 \draw(0.0,4.4) node[inner sep=0pt, opacity=0.9] (img1) {
     \includegraphics[width=4.0cm,height=4.0cm]{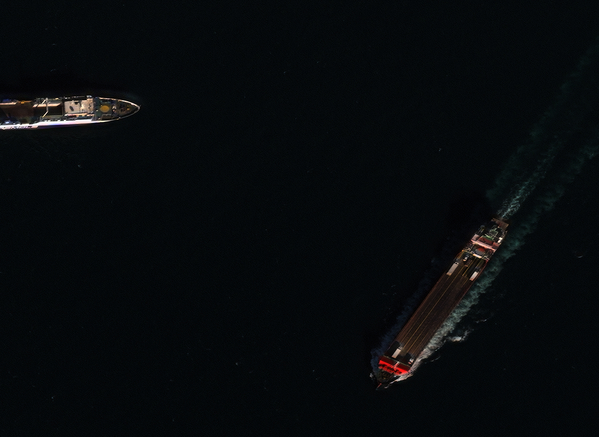}   
  };

  \draw[white=0.9,thick] (-1.95,5.65) rectangle (-1.0,5.15);
  \draw[white,fill=white] (-1.95,5.65) circle (.3ex);
 \draw[white,fill=white] (-1.0,5.15) circle (.3ex);
 \path[->](-1.45,5.85) node[white,very thick, inner sep=1pt]  {\small \textbf{$r_2$}};
 
  \draw[white=0.9,thick] (+0.5,4.0) rectangle (+1.5,2.9);
   \draw[white,fill=white] (+0.5,4.0) circle (.3ex);
 \draw[white,fill=white] (+1.5,2.9) circle (.3ex);
 \path[->](+0.4,3.6) node[white,very thick, inner sep=1pt]  {\small \textbf{$r_3 (w = 0.8)$}};
   
  \draw[white=0.9,thick] (+0.4,4.9) rectangle (+1.8,2.82);
   \draw[white,fill=white] (+0.4,4.9) circle (.3ex);
 \draw[white,fill=white] (+1.8,2.82) circle (.3ex);
  \path[->](+1.0,5.1) node[white,very thick, inner sep=1pt]  {\small \textbf{$r_1 (w = 0.7)$}};
  
  \draw(4.2,4.4) node[inner sep=0pt, opacity=0.9] (img1) {
     \includegraphics[width=4.0cm,height=4.0cm]{figures/crop_2270.jpg}   
  };
  
   \path[->](-0.9,2.1) node[black,very thick,fill=white,fill opacity=0.9, inner sep=-2pt]  {\small $\mathcal{R} = \{r_1, r_2, r_3\}$};
  
   \path[->](3.3,2.1) node[black,very thick,fill=white,fill opacity=0.9, inner sep=1pt]  {\small $\mathcal{R}_1 = \{r_2\} = \bar{r}_1$};
  
  \path[->](3.52,1.7) node[black,very thick,fill=white,fill opacity=0.9, inner sep=1pt]  {\small $\mathcal{R}_2 = \{r_1, r_3\} = \bar{r}_2$};
  
  \path[->](3.23,1.3) node[black,very thick,fill=white,fill opacity=0.9, inner sep=1pt]  {\small $\mathcal{D} = \{\bar{r}_1, \bar{r}_2\}$}; 
     
    \draw[white=0.9,thick] (2.25,5.65) rectangle (3.2,5.15);
  \draw[white,fill=white] (2.25,5.65) circle (.3ex);
 \draw[white,fill=white] (3.2,5.15) circle (.3ex);
 \path[->](+2.8,5.85) node[white,very thick, inner sep=1pt]  {\small \textbf{$\bar{r}_1$}};
  
  \draw[white=0.9,thick] (+4.6,4.5) rectangle (+5.8,2.82);
   \draw[white,fill=white] (+4.6,4.5) circle (.3ex);
 \draw[white,fill=white] (+5.8,2.82) circle (.3ex);
  \path[->](+5.2,4.7) node[white,very thick, inner sep=1pt]  {\small \textbf{$\bar{r}_2$}};
  
\end{tikzpicture}
  \caption{Region merging: on left, three regions were detected; on right, two regions were merged by using the confidence score of both to define the new dimensions.} 
  \label{fig:merging}
\end{figure}

In our framework, the merging algorithm also produces $k$ regions 
\begin{equation}
  D = \{\bar{r}_1, \bar{r}_2, \dots, \bar{r}_k\}
\end{equation}
however, instead of discarding many overlapping regions with high confidence score, we merge the axis-aligned region rectangles within each subset $\mathcal{R}_i$ considering a weighted average criterion, that is, 
\begin{equation}
  {\bar{r}_i}(b) = \frac{\displaystyle \sum_{r \in \mathcal{R}_i} r(w) \times r(b)}{\displaystyle \sum_{r \in \mathcal{R}_i} r(w)} 
  \end{equation}
where $r(w) \times r(b) = r(w) \times (x_1, y_1, x_2, y_2)$. See Figure~\ref{fig:merging}.

\begin{figure}[!htb]
\centering
\begin{tikzpicture}

  \draw(0.0,-1.0) node[inner sep=0pt, opacity=0.9] (img1) {
     \includegraphics[width=4.0cm,height=4.0cm]{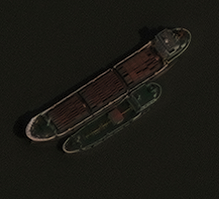} };
  
    
  \draw[white=0.9,thick] (-1.6,0.5) rectangle (+1.6,-1.9);
  
  \draw[white=0.9,thick] (-0.9,-0.6) rectangle (+1.0,-2.1);
  
  

\end{tikzpicture}

  \caption{Two overlapped objects from the same category.} 
  \label{fig:region_overlap}
\end{figure}

A key aspect in our merging algorithm is the \textit{intersection over union} metric to compute the region intersection, namely
\begin{equation}
 \textrm{IoU}({r_i}(b), {r_j}(b)) = \frac{\textrm{area\;}({r_i}(b) \cap {r_j}(b))}{\textrm{area\;}({r_i}(b) \cup {r_j}(b))} > \sigma
\end{equation}
The IoU metric, as opposed to the \emph{intersection score},  takes into account the total area of both regions. This is particularly interesting for the xView challenge, where in many cases a significant intersection of objects does not mean that they should be merged, see Figure~\ref{fig:region_overlap}.

\section{Experiments}\label{sec:results}

\subsection{Dataset}

The xView dataset, as detailed by Darius et al.~\cite{xview2018}, contains 1,413 high-resolution images, with image area ranging from $2564 \times 2576$ to $3187 \times 4994$ pixels, spanning approximately one million objects from 60 categories. The dataset was split into training, evaluation and testing subsets, as shown in Table~\ref{tab:xview_dataset}. 

\begin{table}[!ht]
 \renewcommand*{\arraystretch}{1.2}
 \centering
 \caption{The xView dataset. The evaluation and testing ground-truths regions were not released.}
 \setlength\tabcolsep{2.9pt}
 \begin{tabular}{lrllllll}
  \toprule
  & {\bf \#images} & {\bf \#regions} &  {\bf \#small} &  {\bf \#medium} &  {\bf \#large}  &  {\bf \#common} &  {\bf \#rare} \\
  \midrule
  Train       & 847  & 601,345 & 256,793 & 333,406 & 11,659 & 595,149 & 6,709\\
  Eval.        & 282  & 200,291 & --- & --- & --- & --- & ---\\
  Test         & 284  & --- & --- & --- & --- & --- & ---\\
  \bottomrule
  \textbf{Total} & 1,413  & 800,636 & --- & --- & --- & --- & ---  \\
  \bottomrule
 \end{tabular}
 \label{tab:xview_dataset}
\end{table}

The xView contest divided the dataset into three classes of object size in order to report the algorithm's performance:
\begin{itemize}
 \item \textbf{Small} objects: passenger-vehicle, small-car, bus, pickup-truck, utility-truck, truck, cargo-truck, truck-tractor, trailer, truck-tractor-w-flatbed-trailer, crane-truck, motorboat, dump-truck, scraper-tractor, front-loader-bulldozer, excavator, cement-mixer, ground-grader and shipping-container.
 \item \textbf{Medium} objects: fixed-wing-aircraft, small-aircraft, helicopter, truck-tractor-w-box-trailer, truck-tractor-w-liquid-tank, railway-vehicle, passenger-car, cargo-container-car, flat-car, tank-car, locomotive, sailboat, tugboat, fishing-vessel, yacht, engineering-vehicle, reach-stacker, mobile-crane, haul-truck, hut-tent, shed, building, damaged-building, helipad, storage-tank, pylon and tower.
 \item \textbf{Large} objects: passenger-cargo-plane, maritime-vessel, barge, ferry, container-ship, oil-tanker, tower-crane, container-crane, straddle-carrier, aircraft-hangar, facility, construction-site, vehicle-lot and shipping-container-lot.
\end{itemize}
Another division of the same objects considered their presence in the dataset:
\begin{itemize}
 \item \textbf{Rare} objects: fixed-wing-aircraft, small-aircraft, helicopter, truck-tractor-w-liquid-tank, crane-truck, railway-vehicle, flat-car, tank-car, locomotive, maritime-vessel, sailboat, tugboat, barge, ferry, yacht, container-ship, oil-tanker, engineering-vehicle, tower-crane, container-crane, reach-stacker, straddle-carrier, mobile-crane, haul-truck, scraper-tractor, cement-mixer, ground-grader, aircraft-hangar, helipad, pylon and tower;
 \item \textbf{Common} objects: passenger-cargo-plane, passenger-vehicle, small-car, bus, pickup-truck, utility-truck, truck, cargo-truck, truck-tractor-w-box-trailer, truck-tractor, trailer, truck-tractor-w-flatbed-trailer, passenger-car, cargo-container-car, motorboat, fishing-vessel, dump-truck, front-loader-bulldozer, excavator, hut-tent, shed, building, damaged-building, facility, construction-site, vehicle-lot, storage-tank, shipping-container-lot and shipping-container. 
\end{itemize}

\subsection{Hardware Time restrictions}

The participants needed to submit their solutions in the xView system by using a docker container with the inference source code, trained models and required packages. The solution also ran inferences for the validation set by respecting the following hardware limitations:
\begin{itemize}
  \item The inference for an input image must be completed in less than 40 minutes.
  \item Evaluating the entire validation set should not take more than 72 hours.
  \item The inference process need to use a cluster of Central Processing Units (CPUs), with a memory limit of 8 GB.   
\end{itemize}
  
The xView challenge used the validation set, with known images but unknown regions labels, to provisionally rank the participants solutions. The final ranking was determined by the performances in a sequestered test dataset.

\subsection{Settings}~\label{sec:settings}

Table~\ref{tab:parameters_table} shows the parameter configuration for each pipeline in terms of image scaling, region overlap (horizontal and vertical directions), the confidence threshold for region filtering, the baseline trained model~\cite{xview2018} used for inference and the objects to be detected defined by the category size.

\begin{table}[!htb]
\renewcommand*{\arraystretch}{1.2}
\setlength\tabcolsep{6pt}
\caption{Parameter settings}
\label{tab:parameters_table}
\begin{center}
\begin{tabular}{|c||l|c|}
\hline
& Parameters & Value\\
\hline \hline
\multirow{4}{*}{Pipeline 1} & Image scaling & $1.0$ \\ 
 & Region overlap & 0 pixels (no-overlap) \\ 
& Confidence threshold & 0.15 \\ 
& Classification model & Vanilla (SR)\\ 
& Object of interest (by size) & Small and medium \\
\hline \hline 
\multirow{4}{*}{Pipeline 2} & Image scaling & $1.3$ \\ 
 & Region overlap & 0 pixels (no-overlap) \\ 
& Confidence threshold & 0.06 \\ 
& Classification model & Vanilla (SR)\\ 
& Object of interest (by size) & Small and medium \\
\hline \hline 
\multirow{4}{*}{Pipeline 3} & Image scaling & $0.7$ \\ 
 & Region overlap & 100 pixels \\ 
& Confidence threshold & 0.5 \\ 
& Classification model & Multires (MR)\\ 
& Object of interest (by size) & Medium and large \\
\hline \hline 
\multirow{4}{*}{Pipeline 4} & Image scaling & $1.0$ \\ 
 & Region overlap & 100 pixels \\ 
& Confidence threshold & 0.06 \\ 
& Classification model & Multires (MR)\\ 
& Object of interest (by size) & Small, medium and large \\
\hline \hline 
\multirow{4}{*}{Pipeline 5} & Image scaling & $0.6$ \\ 
 & Region overlap & 0 pixels (no-overlap) \\ 
& Confidence threshold & 0.06 \\ 
& Classification model & Multires (MR)\\ 
& Object of interest (by size) & Large \\
\hline 
\end{tabular}
\end{center}
\end{table}

\subsection{Results}

The primary quantitative criteria used by the xView challenge for ranking purposes, is the interpolated mean average precision (mAP) metric, detailed by Henderson and Ferrari~\cite{DBLP:journals/corr/HendersonF16}. Informally, this metric sort the predicted rectangles by the confidence score, in descending order, and then, if the intersection over union (IOU) metric is above 0.5 for a pair of predicted and groundtruth regions, then we have a true positive, otherwise, the matching is considered  a false positive --- undetected groundtruth regions are considered false negatives. The mAP performances for xView were computed by the challenge submission system, as shown in Table~\ref{table:xview_performance}.

\begin{table}[!htb]
  \centering
  \renewcommand{\arraystretch}{2.0}%
  \setlength\tabcolsep{7pt}
  \caption{The mean average precision (mAP) score for the xView validation subset.}
  \begin{tabular}{|l||c|} \cline{2-2}
   \multicolumn{1}{c||}{}  & Mean average precision (mAP) \\ \cline{1-2}
   \textbf{Proposed} framework & \textbf{29.88}  \\ \hline   
   Vanilla (SR)    &  20.87  \\ 
   Multires (MR)   &  18.14  \\ \hline 
  \end{tabular}
  \label{table:xview_performance}  
\end{table}

\begin{figure}[!htb]
   \centering
   \includegraphics[scale=0.48]{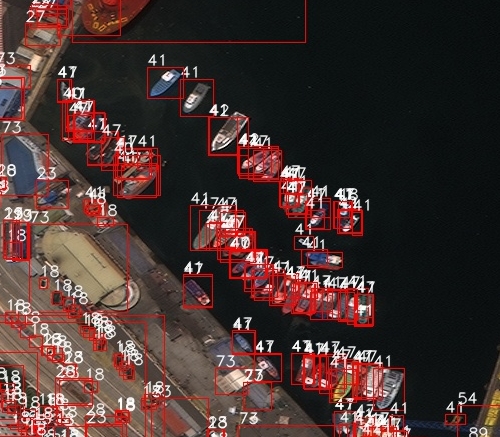}
   \caption{Example of object detection with the proposed framework.}
   \label{fig:detection} 
\end{figure}

\section{Discussion}~\label{sec:discussion}


We outline here some ideas that we tried, but did not result in good performance. Experiments with other popular deep learning approaches for object detection and classifications such as Faster-RCNN, SSD, RetinaNet and You Only Look Once version 3 (YOLOv3) \cite{redmon2018yolov3}, resulted in only meager performance increases. 

Strategies that typically go with deep learning approaches such as optimization techniques, data augmentation, drop out, using different feature extraction architectures and varying learning rates were applied with limited success. 

The predictions before and after post-processing stage were significantly high, for example, the number of predictions made on a subset of the training data was at least five times more than the number of ground truth in which a good number were false detections. Reducing this number using by thresholding led to significant increase in accuracies, however, a better approach might lead to a higher gain.

Input augmentation during inference also appeared to be a technique that could enable better detection and classification, for the challenge we successfully utilized zoom, however, horizontal flips was not as effective and probably other forms of augmentation may yield better results.

\section{Ongoing Research}~\label{sec:future_work}

\begin{figure}[!htb]
   \centering
   \includegraphics[scale=0.35]{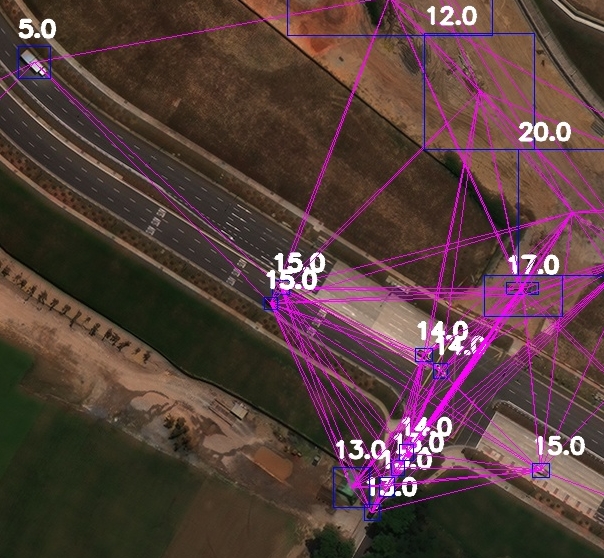}
   \caption{Topological spatial relationships of objects in remote sensing.}
   \label{fig:graph} 
\end{figure}

A rich source of structural information, that could be exploited to improve the classification, is the topological spatial relationship inherent to many classes of objects in the aerial imagery context. 
A \emph{graph} formed by such relationships is shown in Figure~\ref{fig:graph}, where the vertices are the object regions and the edges represent the shortest distance.

\begin{figure}[!htb]
  \begin{tikzpicture}
  \draw(1.2,2.0) node[inner sep=0pt] (img1) {
     \includegraphics[width=0.5\textwidth]{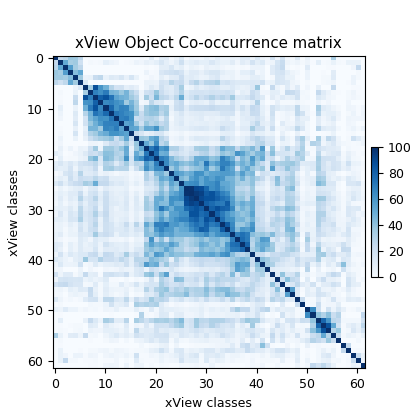}
  };
   \path[->](-0.7,4.95) node[]  {\normalsize Group 1};
  \draw[very thick,red] (-1.8,4.95) circle (0.43cm);

  \path[->](+0.4,4.1) node[]  {\normalsize Group 2};
  \draw[very thick,red] (-0.9,4.1) circle (0.7cm);

  \path[->](+1.19,3.1) node[]  {\normalsize Group 3};
  \draw[very thick,red] (+0.1,3.1) circle (0.43cm);

  \path[->](2.58,2.0) node[]  {\normalsize Group 4};
  \draw[very thick,red] (+1.18,2.0) circle (0.7cm);

  \path[->](2.8,1.32) node[]  {\normalsize Group 5};
  \draw[very thick,red] (+1.8,1.32) circle (0.3cm);

  \path[->](4.6,-0.45) node[]  {\normalsize Group 6};
  \draw[very thick,red] (+3.6,-0.45) circle (0.32cm);

  \draw(1.2,-8.0) node[inner sep=0pt] (img1) {
     \includegraphics[width=0.5\textwidth]{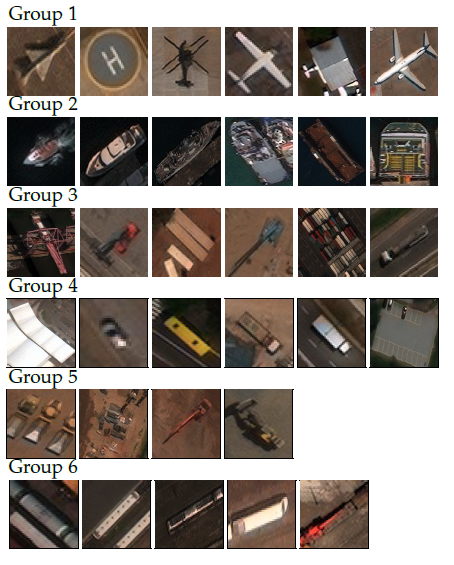}
  };
  \end{tikzpicture}
   \caption{Spatial context: the co-occurrence matrix and six clusters of objects. \textbf{Group 1}: fixed-wing aircraft, helipad, helicopter, small-aircraft, aircraft-hangar and passenger cargo-plane.
  \textbf{Group 2}: sailboat, fishing-vessel, motorboat, yacht, maritime-vessel, tugboat", barge, ferry, container-ship, oil-tanker;
  \textbf{Group 3}: container-crane, reach-stacker, shipping-container, mobile-crane, shipping-container-lot and truck-tractor-w-flatbed-trailer; \textbf{Group 4} building, small-car, bus, truck, cargo-truck, vehicle-lot, utility-truck, truck-tractor-w-box-trailer; \textbf{Group5}: dump-truck, construction-site, excavator and front-loader-bulldozer; and \textbf{Group 6}: tank-car, cargo-container-car, passenger-car, locomotive.}
   \label{fig:co-occurrence}
\end{figure}
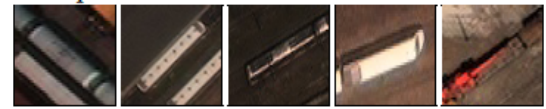

The spatial context for the xView benchmark is depicted by the co-occurrence matrix, shown in Figure~\ref{fig:co-occurrence}, where one can easily see some clusters of objects that are part of the same scene shot.

We are working to use this spatial graph to filter out the false positives regions. Specifically, to verify if such neighborhood relations make sense in the real world one can use the training set or semantic networks such as ConceptNet~\cite{Liu2004} or geographical statistics from OpenStreetMap~\cite{4653466} and change the classifications based on this prior knowledge.

\section{Acknowledgement}

We appreciate the discussions with our colleagues Sathyanarayanan Aakur, Mauricio Pamplona Segundo, and Daniel Sawyer as we were developing this approach. 


\bibliographystyle{IEEEtran}
\bibliography{bibliography.bib}
\end{document}